%% file: main.tex
\pdfoutput=1
\documentclass[letterpaper, 10 pt, conference]{IEEEtran}  %

\IEEEoverridecommandlockouts                              %

\usepackage[bookmarks=false]{hyperref}
\usepackage{graphics} %
\usepackage{epsfig} %
\usepackage{mathptmx} %
\usepackage{times} %
\usepackage{amsmath} %
\usepackage{amssymb}  %

\usepackage[T1]{fontenc}
\usepackage[utf8]{inputenc}
\usepackage{amsfonts}
\usepackage{booktabs}
\usepackage{siunitx}
\usepackage{caption} 
\usepackage{booktabs}

\usepackage{array}    %

\usepackage{graphicx}
\usepackage{algorithm, algpseudocode}
\usepackage{float}
\usepackage[percent]{overpic}
\usepackage{caption}
\DeclareGraphicsExtensions{.jpeg,.png}

\usepackage[font=small,skip=0pt]{subcaption}

\usepackage{array, tabularx, makecell, booktabs}%

\usepackage{tikz}

\usepackage{geometry}
\usepackage{comment}

\usepackage{hyperref}
\hypersetup{
    colorlinks=true,
    linktoc=all,
    citecolor=black,
    filecolor=black,
    linkcolor=black,
    urlcolor=black
}

\usepackage{footmisc}

\usepackage{xcolor}

\usepackage{parskip}

\usepackage{listings}

\title{\LARGE \bf
Demonstrating Arena 5.0: A Photorealistic ROS2 Simulation Framework for Developing and Benchmarking Social Navigation
}

\author{ 
    Volodymyr Shcherbyna$^{2,*}$, 
    Linh K{\"a}stner$^{1,2,*}$,
    Duc Anh Do$^{1}$, 
    Hoang Tung$^{3}$,\\
    Huu Giang Nguyen$^{2}$,
    Maximilian Ho-Kyoung Schreff$^{3}$, 
    Tim Seeger$^{2}$,\\
    Eva Wiese$^{2}$,
    Ahmed Martban$^{2}$,
    Huajian Zeng$^{3}$,
    An Tran$^{1}$,
    Nguyen Quoc Hung$^{2}$,\\
    Jonas Kreutz$^{2}$,
    Vu Thanh Lam$^{2}$,
    Ton Manh Kien$^{2}$,
    Harold Soh$^{1}$\\
    {\small $^{*}$Equal Contribution}\\
    {\small $^{1}$National University of Singapore (NUS), Singapore} \\
    {\small $^{2}$Technical University Berlin (TUB), Germany} \\
    {\small $^{3}$Technical University Munich (TUM), Germany}
}

\usepackage{makecell}

\usepackage{makecell}

\begin{document}

\maketitle

\begingroup
\renewcommand\thefootnote{}
\footnotetext{\hspace*{-2.2em}This research / project is supported by A*STAR under its National Robotics Programme (NRP) (Award M23NBK0053). This research is supported in part by the National Research Foundation (NRF), Singapore and DSO National Laboratories under the AI Singapore Program (Award Number: AISG2-RP-2020-017).}
\endgroup

\input{ieeeconf/content/0-abstract}
\input{ieeeconf/content/1-introduction}
\input{ieeeconf/content/2-Related-Works}
\input{ieeeconf/content/3-methodology}
\input{ieeeconf/content/4-evaluations}
\input{ieeeconf/content/5-conclusion}

\typeout{}
\bibliographystyle{IEEEtran}
\bibliography{main}
\clearpage

\input{ieeeconf/content/6-appendix}

\addtolength{\textheight}{-1cm} 

\end{document}

%% file: ieeeconf/content/0-abstract.tex
\begin{abstract}
\noindent
Building upon the foundations laid by our previous work, this paper introduces \textbf{Arena 5.0}, the fifth iteration of our framework for robotics social navigation development and benchmarking. Arena 5.0 provides three main contributions: 1) The complete integration of NVIDIA Isaac Gym, enabling photorealistic simulations and more efficient training. It seamlessly incorporates Isaac Gym into the Arena platform, allowing the use of existing modules such as randomized environment generation, evaluation tools, ROS2 support, and the integration of planners, robot models, and APIs within Isaac Gym. 2) A comprehensive benchmark of state-of-the-art social navigation strategies, evaluated on a diverse set of generated and customized worlds and scenarios of varying difficulty levels. These benchmarks provide a detailed assessment of navigation planners using a wide range of social navigation metrics. 3) Extensive scenario generation and task planning modules for improved and customizable generation of social navigation scenarios, such as emergency and rescue situations. The platform's performance was evaluated by generating the aforementioned benchmark and through a comprehensive user study, demonstrating significant improvements in usability and efficiency compared to previous versions. Arena 5.0 is open source and available at \href{https://github.com/Arena-Rosnav}{\textcolor{blue}{https://github.com/Arena-Rosnav}}.

\end{abstract}

%% file: ieeeconf/content/1-introduction.tex
\section{Introduction}
\begin{figure}[!htp]
    \centering
    \includegraphics[width=0.99\linewidth]{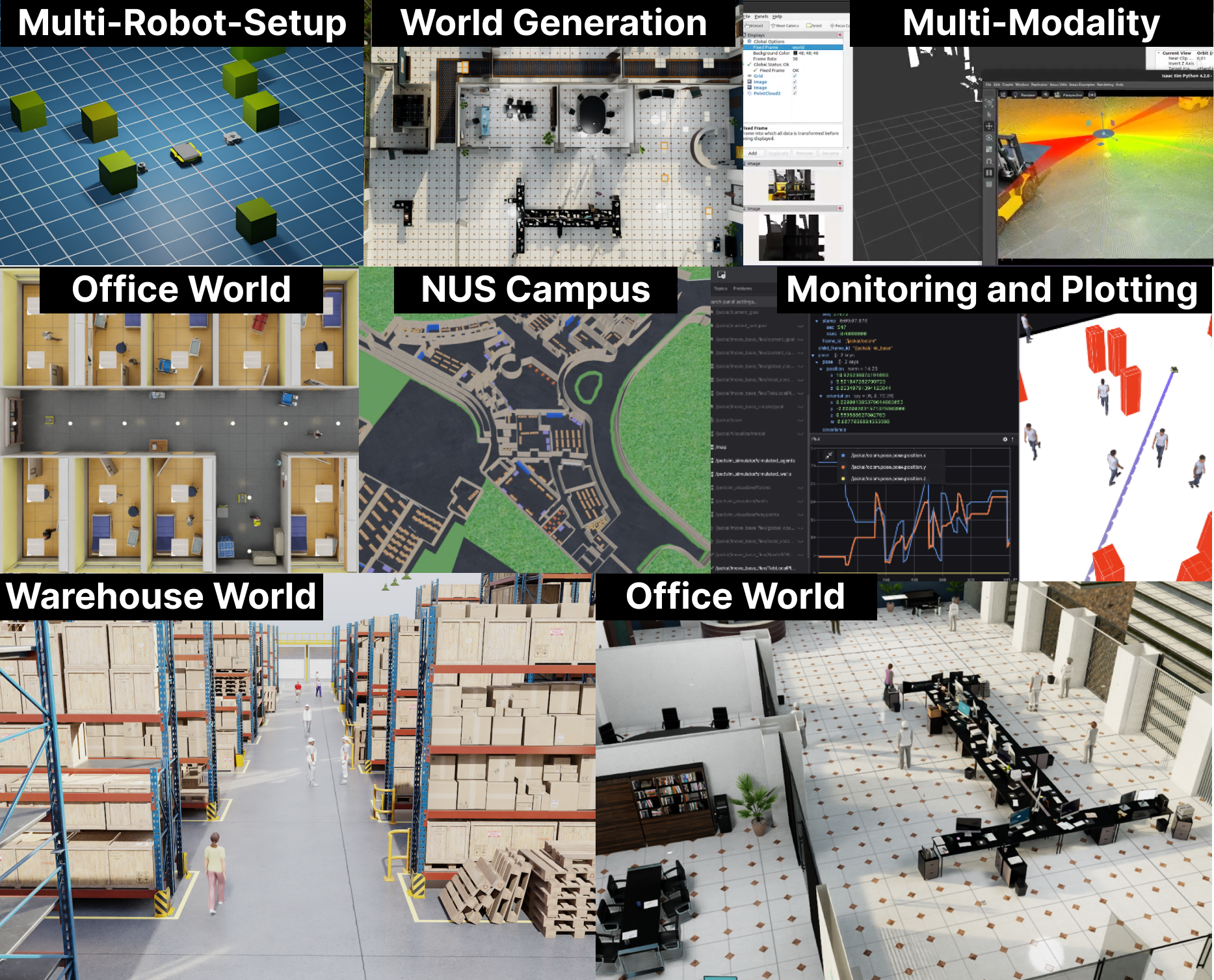}
    \caption{Sample scenes from the Arena 5.0 platform, which provides tools to develop social navigation approaches in highly dynamic and crowded environments. It focuses on social navigation and provides a number of modules to achieve realistic simulation of human-centric environments, developing and testing navigation algorithms on various robotic systems and setups, and simplified extension with new modules.}
    \label{intro}
\end{figure}
\noindent Social navigation has gained increasing importance across various sectors such as healthcare, logistics, and assistive robotics. Despite its growing adoption, navigating human-centric environments remains a significant challenge. A major concern is bridging the gap between simulation-based research and industry-grade applications (sim2real) \cite{mirsky2024conflict}. Consequently, there is a growing demand for realistic and comprehensive simulation platforms, as emerging approaches must be thoroughly tested and benchmarked prior to real-world deployment. Common issues include non-reproducible research findings and highly specialized outcomes that are difficult to generalize for practical applications. Moreover, many existing platforms continue to rely on legacy software (e.g., ROS1), complicating the transition from simulation to reality and further widening the sim2real gap. This reliance impedes progress and underscores the necessity for updated simulation frameworks that better facilitate technology transfer to industry-grade systems \cite{singamaneni2024survey}.
\begin{figure*}[!ht]
    \centering
    \includegraphics[width=0.99\linewidth]{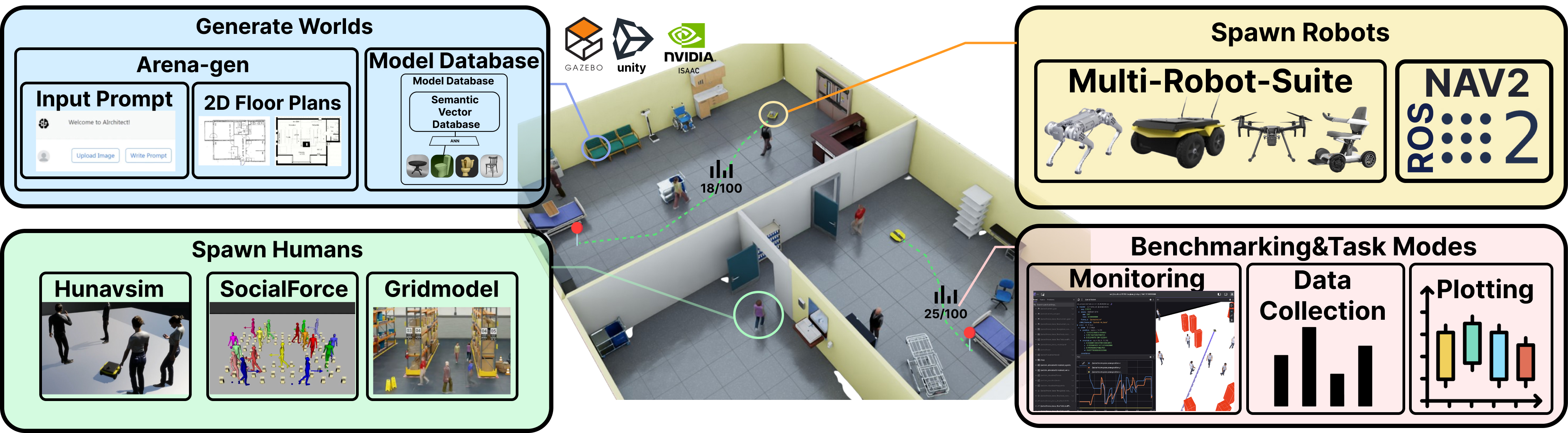}
    \caption{Data flow of the \emph{Generation Stage} and the \emph{Population Stage}. The Generation Stage combines multiple SotA technologies to process text inputs into a floor plan image and room asset locations. 3DSGs are used as an intermediate data structure to divide the problem into a text transformation task solvable by an LLM, and a graph transformation task solvable by a spatial GNN. The \emph{Population Stage} populates the floor plan's asset zones with 3D models by employing the \emph{Asset Placer}. A pre-built semantic vector \emph{Model Database} is queried for a related model, which is arranged into the zone by a \emph{Fitter} algorithm. After a final post-processing step, the end result is a finished environment consisting of 3D walls and models.}
    \label{img:gen-dataflow}
\end{figure*}
\\\noindent
Most open-source benchmarking platforms concentrate on specific navigation aspects without addressing broader concerns such as real-robot integration, deployment, and practical applicability. Owing to varied setups—ranging from in-house simulators to proof-of-concept demonstrations—numerous approaches are overly simplified and challenging to replicate in real-world scenarios. Additionally, many available simulation environments either cater to specialized tasks (e.g., learning-based methods) or offer limited functionality. In contrast, sophisticated state-of-the-art simulators such as Isaac Sim present substantial usability challenges, are not fully open source, and consequently have a restricted user base \cite{singamaneni2024survey, mirsky2024conflict, francis2023principles}. 
\\\noindent
To address these issues, we introduced world generation capabilities and social interaction modules in Arena 3, followed by a platform-wide migration to ROS2 and extended world-generation functionalities in Arena 4.0. Building on these foundations, Arena 5.0 integrates Isaac Gym to offer photorealistic simulations, thereby aligning with both research and industry requirements. By providing an enhanced world and scenario generation process, as well as a direct link to Isaac Gym, Arena 5.0 enables more realistic and comprehensive evaluations, bridging a critical gap in social navigation research and deployment.
\\\noindent
The main contributions of this work are as follows:
\begin{itemize}
    \item \textbf{Full Integration of Isaac Gym into the Arena Framework:} This includes a comprehensive ROS integration of Isaac Gym, facilitating seamless inter-operation with ROS-based systems. This integration aims to leverage Isaac Gym's advanced simulation capabilities within the Arena framework to enhance the development and testing of autonomous robots in realistic environments.
    
    \item \textbf{Benchmarking:} Establishment of rigorous benchmarks for evaluating the performance of social navigation algorithms. This includes metrics for assessing accuracy, efficiency, and real-time performance of robots within simulated human-centric environments.
    
    \item \textbf{Automated and Customizable Scenario Generation Pipelines:} Development of flexible scenario generation pipelines that allow researchers to automatically create diverse and complex environments. This feature supports customization to address specific research needs and scenarios, thus accelerating the testing and refinement of navigation algorithms.
\end{itemize}

\color{black}

%% file: ieeeconf/content/2-Related-Works.tex
\begin{figure*}[!htp]
    \centering
    \includegraphics[width=0.99\linewidth]{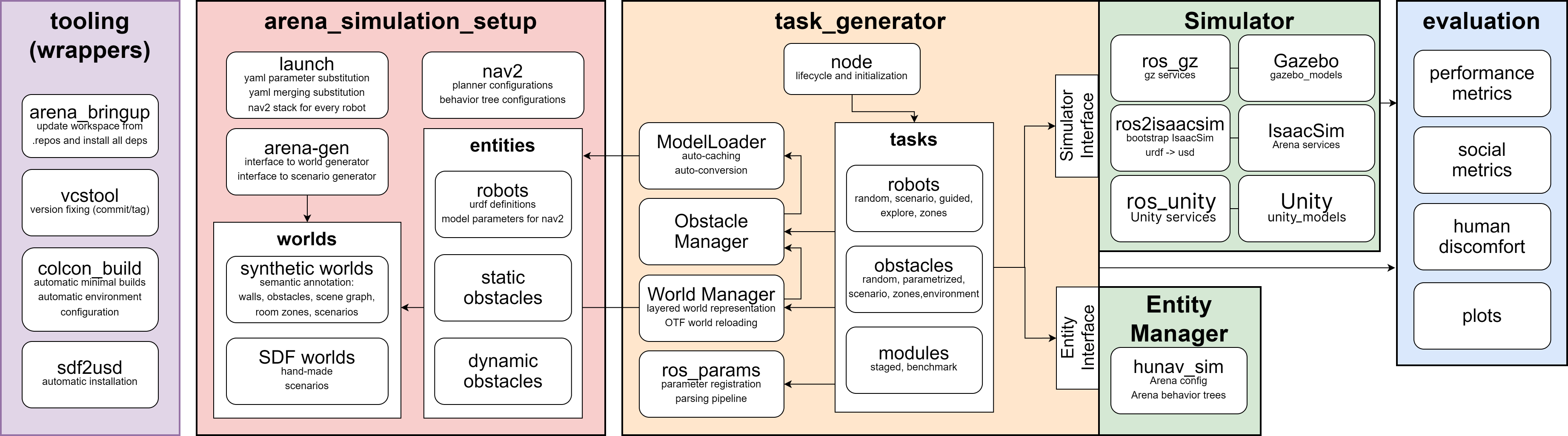}
    \caption{\emph{System Design of Arena 5.0}: Our central modules are \texttt{arena\_simulation\_setup} and \texttt{task\_generator} and, which provide interfaces for loading worlds, defining world semantics, providing scenarios and benchmarking configurations, managing robot and non-robot entities, interacting with the parameter server, interfacing the navigation stack, photorealistic simulators, pedestrian simulators. The Python-only modules are highly extensible and provide a convenient API that simplifies interactions with both Arena and the ROS2 ecosystem. We provide additional tooling in the form of installers and build/version management tools, building on top of widely used technologies \texttt{colcon} and \texttt{vcstool}. Our peripheral modules \texttt{Arena-Gen} and \texttt{Arena-Evaluation} are installed alongside the core modules and are interfaced directly and implicitly based on the user's intentions. The lifecycle management of the system is tied together in our separate centralized module \texttt{arena\_bringup}.}
    \label{img:system-design}
\end{figure*}
\section{Related Works}
\noindent Social navigation is increasingly critical for daily tasks, and a variety of research works have been conducted for the development, training, and testing of navigation approaches in social environments \cite{wang2024srlm}, \cite{raj2024rethinking}, \cite{chen2024socially}, \cite{song2024socially}. However, although many social navigation approaches continue to emerge, their reproducibility, applicability, and validity are often hard to assess due to highly specific settings, simulators, labs, and installation setups. In a recent studies, by Francis et al. and Singamaneni et al. \cite{singamaneni2023towards}, \cite{singamaneni2024survey} the lack of a unified benchmark for social navigation approaches is highlighted as one of the main bottlenecks in social navigation robotics research. 
Navigation benchmarks such as Bench-MR by Heiden et al. \cite{bench_mr}, Robobench by Weisz et al. \cite{robobench}, CommonRoad by Althoff et al. \cite{althoff2017commonroad}, and the Benchmarking Suite by Moll et al. \cite{moll2015benchmarking} focus solely on 2D simulation and only consider static environments.
Other benchmarks such as SocNavBench \cite{biswas2022socnavbench} by Biswas et al., DynaBARN by Nair et al. \cite{nair2022dynabarn}, or SocialGym by Holtz et al. \cite{holtz2021socialgym} employ dynamic environments but contain simplistic and limited navigation approaches for proof of concepts. Other social navigation benchmarking platforms include SEAN and SEAN2.0 by Tsoi et al. \cite{tsoi2020sean, tsoi2022sean}, HuNavSim by Perez-Higueras et al. \cite{perezhunavsim}, MRPB 1.0 by Wen et al. \cite{mrpb}, SRPB by Karwowski et al. \cite{karwowski2023srpb}, or the Social Evaluation Platform by Gao et al. \cite{gao2022evaluation}. However, they often face limitations in scenario diversity or robot variety. Furthermore, these environments are frequently constrained by their simulation capabilities or the scenarios they offer. Additionally, they do not cover recent approaches utilizing other sensor modalities such as rgbd, sonar, imu sensors, and often utilize legacy software frameworks. Other social navigation benchmarks focus on incorporating multi-agent reinforcement learning, such as SAMARL by Wang et al. \cite{wang2024multi} or Socialgym2.0 \cite{sprague2023socialgym}. However, they are limited to 2D simplistic simulations and showcase proof-of-concepts approaches. As industry is shifting towards ROS2 usage, and with the advent of increased computational power, photorealistic simulators and more sensor modalities are often required to employ sophisticated approaches. Notably, Isaac Gym \cite{makoviychuk2isaac} by Makoviychuk et al. has been making strides in generating diverse, complex, dynamic environments and has also been adopted by numerous laboratories. However, due to its recent release, the documentation and usability remain challenging and are primarily accessible to experts in the field. Additionally, it is not open source, limiting its availability to a wider audience. To address aforementioned issues, we propose Arena 5.0, which offers a variety of different simulation environments, sensor modalities, a comprehensive suite of robots and planners, as well as modules for complex world and scenario generation in social settings making it a useful platform for social navigation robotics research.

%% file: ieeeconf/content/3-methodology.tex
\begin{figure*}[!htp]
    \centering
    \includegraphics[width=0.99\linewidth]{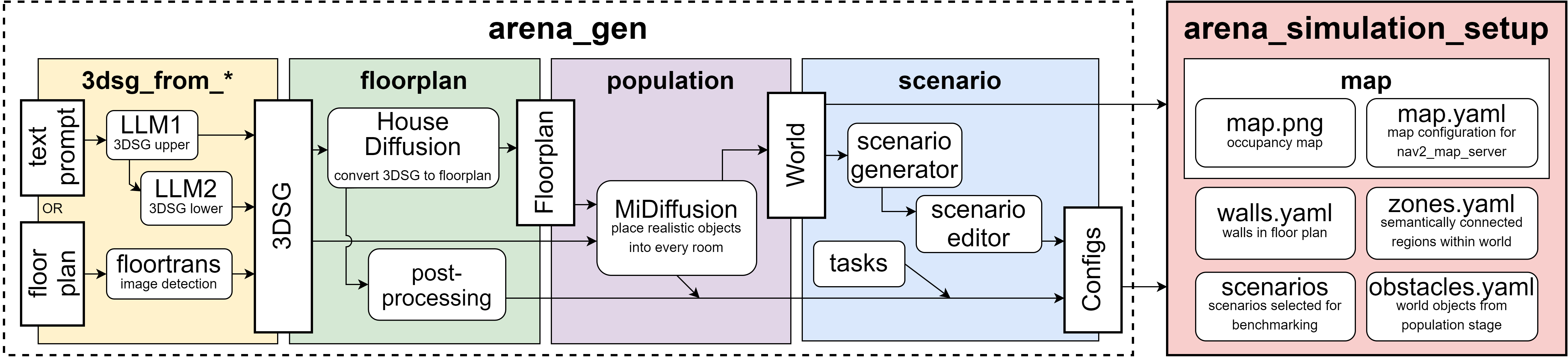}
    \caption{\emph{System Design and Data Flow of arena\_gen}: (1) a user prompt (text or floorplan image) is converted into a 3DSG and edited by the user; (2) a floorplan is generated by the HouseDiffusion~\cite{housediffusion} and post-processed; (3) objects are placed into the rooms and realistically arranged by the MiDiffusion network, creating a world; (4) a scenario is generated and edited by the user, selecting specific tasks for that world; and finally exported to arena\_simulation\_setup for use in training and evaluation in the rest of the Arena ecosystem. The world consists of a classic map definition, as well as additional structural and semantic annotations used to load the world into a simulator and construct semantically relevant tasks.}
    \label{img:arena_gen}
\end{figure*}
\section{Overview of Arena 5.0}
\noindent Building on our previous works, Arena 5.0 provides a modular software stack designed to support the development and benchmarking of social navigation methods. \autoref{img:system-design} illustrates the core concepts and modules of the Arena platform, focusing on the automated generation of worlds where humans and robots are spawned to create social scenarios. Researchers can then utilize Arena’s benchmarking and task planning modules to conduct thorough testing and validation of their approaches.
\\\noindent
Arena 5.0 introduces three key contributions: (1) the integration of Isaac Gym, (2) a comprehensive benchmarking suite for state-of-the-art social navigation methods incorporating multi-modal sensor data across various robotic platforms, and (3) novel world and scenario generation modules that employ generative AI and Large Language Models (LLMs) to facilitate the intuitive creation of social scenarios and environments.

\subsection{System Design of Arena 5.0}
\autoref{img:system-design} provides an overview of the core concepts and functionalities of Arena 5.0. The platform comprises various modules, most crucially those that generate social environments within a diverse set of 3D photorealistic environments, such as Unity and Isaac Gym. Users have numerous options to create these worlds, including capabilities to spawn humans and specify their behaviors. This flexibility enables the simulation of specific scenarios, such as emergency situations, enhancing the realism and applicability of the research.
\\\noindent
The system design is detailed subsequently, where the connections between each module are highlighted, illustrating the comprehensive integration within the platform. The most novel and significant module is the integration of Isaac Gym. This addition required several adaptations and new modules to fully leverage the capabilities offered by the Arena framework. These enhancements enable users to create world scenarios, conduct benchmarking, and run planners using Isaac Sim as a simulator, thereby expanding the platform's utility and effectiveness in realistic simulations.

\subsection{Isaac Gym Integration}
\noindent
We have developed a wrapper around Isaac Gym to enhance its accessibility and integration within the Arena framework. Despite its widespread adoption across many institutions, Isaac Gym presents significant barriers to entry, primarily due to a lack of comprehensive documentation, tutorials, and support. By embedding Isaac Gym into the Arena framework, we significantly lower these hurdles, facilitating a more user-friendly environment for researchers and developers.
\\\noindent
This integration allows users to effortlessly generate social navigation scenarios that exhibit photorealistic behaviors. The Arena framework leverages Isaac Gym’s advanced simulation capabilities, enabling the creation of complex interactive environments where robotic systems can be tested under conditions that closely mimic real-world settings. This approach not only improves the utility of Isaac Gym but also expands its applicability to a broader range of social navigation research scenarios, thereby contributing to the advancement of human-robot interaction studies.

\subsubsection{Pedestrian Simulation}
\noindent In our developments, we utilize HuNavSim and have fine-tuned other social force models to be compatible with the latest versions of simulators, including Gazebo v2 and Isaac Sim. Moreover, we have incorporated human interactions among pedestrians to render scenarios more realistic and applicable to real-world social navigation contexts.
\\\noindent
Photorealistic visuals are a key feature of our platform, achieved through Isaac Sim's extensive model repository, which includes models representing a diverse range of ages and roles, such as office workers and construction workers. When Arena spawns pedestrians, it selects appropriate models to match the scenario requirements. Pedestrian animations are realistically interpolated, with smooth transitions between small movements based on each evaluation of the Social Force Model (SFM). Furthermore, Arena dynamically switches animations on-the-fly according to the pedestrian's behavior state, enhancing the realism of the simulation.
\\\noindent
To support integration and flexibility, an abstract EntityManager interface is implemented. This interface accommodates different pedestrian simulators, such as HuNavSim, and various SFMs, like LightSFM. Behaviors are managed independently by the designated EntityManager, with Arena responsible for forwarding parameters and interpreting behavior trees. This modular approach allows for seamless integration and management of complex behaviors within the simulation environment, paving the way for more detailed and varied social interaction studies in robotic navigation.

\subsection{World and Scenario Generation}
\noindent In Arena 5.0, we offer distinct worlds where randomized environments can be generated across multiple simulation platforms including Gazebo and Isaac Sim. These worlds encompass a variety of settings such as residential areas, hospitals, offices, and warehouses. Our algorithms employ a method of gradual generation, capable of creating dynamic environments that include both static and dynamic obstacles.
\\\noindent
We use the Arena-Gen module introduced in our previous works and extended it with a refined scene graph LLM (\hyperref[sec:llm]{3dsg\_from\_text}), a new diffusion model for generating intricate worlds (\hyperref[sec:population]{population}), as well as the complete integration into the Isaac Gym pipeline. The user can either upload a 2D floorplan or input a natural language prompt that describes an environment. The new system design and data flow is shown in \autoref{img:arena_gen}. We created our own machine-generated dataset of $(\text{prompt},~\text{rooms}_\text{available},~\text{objects}_\text{available}) \to ((\text{rooms}, \text{doorways}),~\text{rooms}\to\text{objects})$, which is used for the training of both LLMs. Our dataset has a total size of $\approx60000$ entries, with some examples shown in \autoref{tab:dataset}.

\begin{figure}[!htp]
    \centering
    \includegraphics[width=0.95\linewidth]{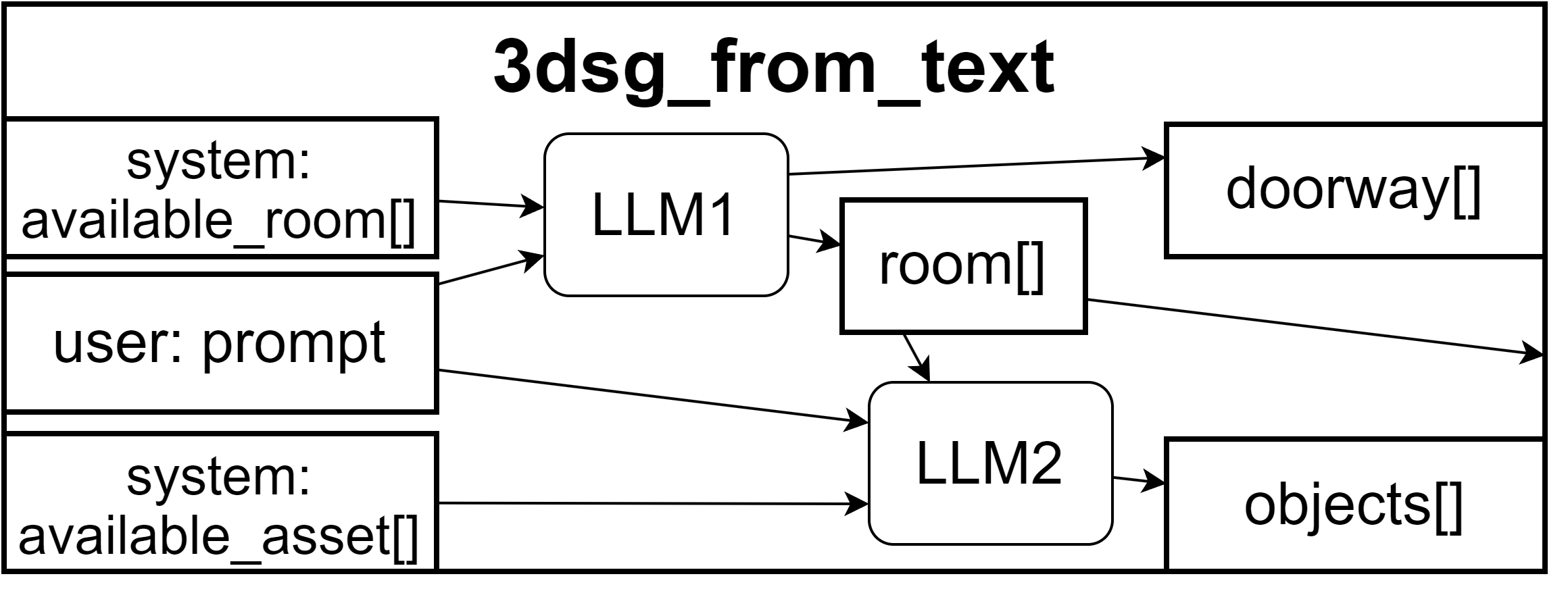}
    \caption{\emph{3dsg\_from\_text architecture}:
        The pipeline architecture features a staggered design that processes different parts of the prompt with different LLMs. \emph{Language Model 1} generates the 3DSG upper half (room and doorway lists) from the user prompt. \emph{Language Model 2} then infers the objects in the rooms from both the user prompt and semantic reasoning capabilities. The lists of possible room and asset types are provided as system prompts and can be changed without re-training.}
    \label{img:3sg_from_text}
\end{figure}

\begin{figure}[!htp]
    \centering
    \includegraphics[width=0.95\linewidth]{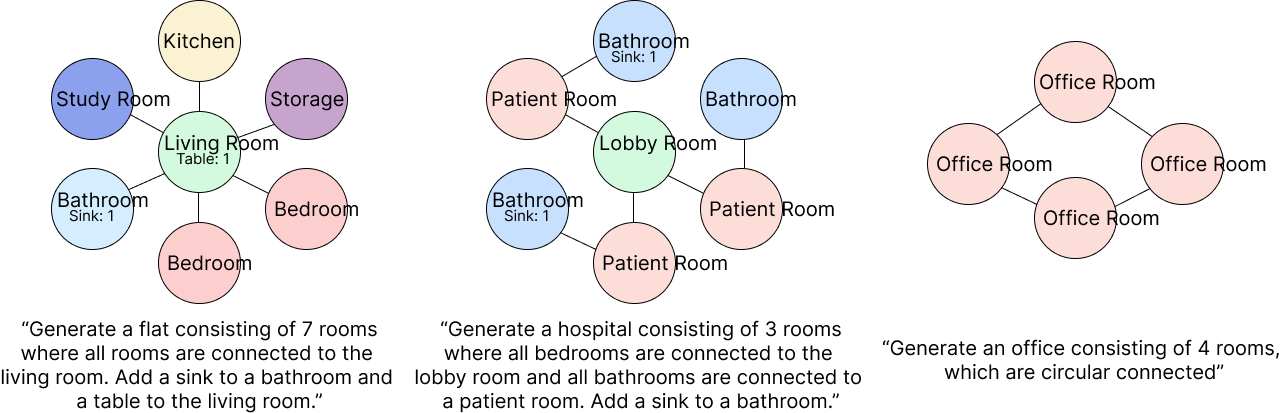}
    \caption{\emph{Prompt Examples}:
        Text prompts and generated graphs to illustrate the geometric and semantic reasoning capabilities of the 3dsg\_from\_text stage. Our architecture is capable of understanding and consistently generating multiple graph types, while also retrieving objects from the prompts and inferring semantically sound objects at the same time. We achieve this with a \emph{divide and conquer} approach by dividing both reasoning tasks between two separate LLMs.}
    \label{img:3dsg_examples}
\end{figure}

\paragraph{3dsg\_from\_text}
\label{sec:llm}
\noindent
We used the pre-trained Google T5-Small~\cite{googlet5} model and fine-tuned it into two sub-models to transform natural language descriptions of room layouts into the respective parts of a 3D Scene Graph. The architecture of this pipeline stage is shown in \autoref{img:3sg_from_text}, with practical examples in \autoref{img:3dsg_examples}.
An overview of the training parameters is shown in \autoref{tab:llm_training}.

\begin{figure}[!htp]
    \centering
    \includegraphics[width=0.95\linewidth]{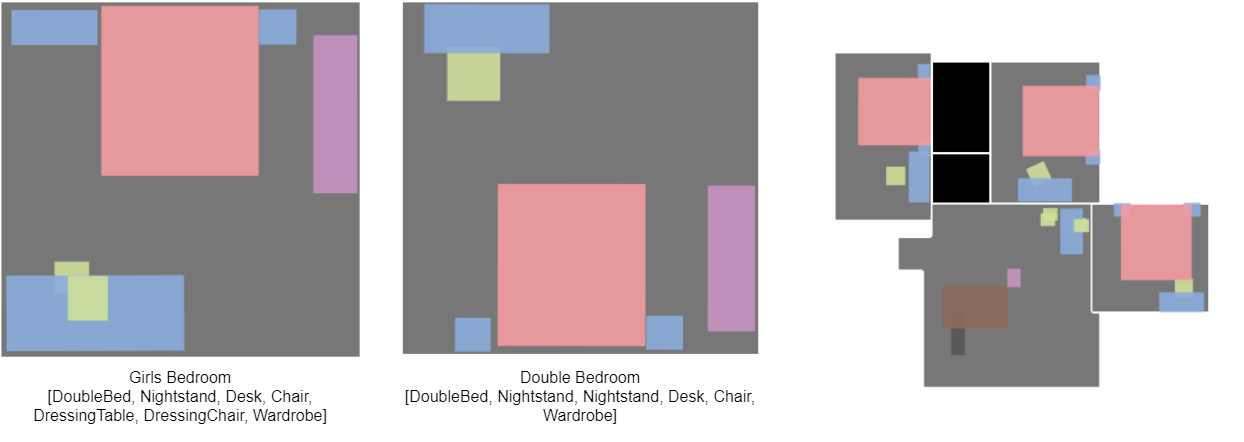}
    \caption{Sample outputs of the population stage.
        \emph{Left}, \emph{Center}: Single-room inference for a given room type and a list of objects. The objects are placed in an intuitive way that allows humans to interact with objects in a realistic way.
        \emph{Right}: Result of population stage applied to an entire residential floorplan. Different room types (bedrooms, living room) are populated alongside each other. The object positioning takes doorways into account and leaves walking space in every room.}
    \label{img:midiffusion}
\end{figure}

\paragraph{Population}
\label{sec:population}
\noindent We employ the mixed diffusion model MiDiffusion~\cite{midiffusion} to arrange a set of objects within each room. Combining the 3D Scene Graph with the output floorplan, the diffusion model successively places the objects into more realistic positions, until a highly realistic placement is finally reached. This process is repeated individually for each room and takes into account the room type. The result is an interactive world that guarantees pathways to exist to reach the objects in each room. This allows the simulation to have more structured human behavior based on human-object interactions. Additionally, robots now navigate a great number of realistic randomized environments, trained on real-world environments, which ensures a smaller sim-to-real gap for human-centric environments.
Example placements illustrating the interactivity are shown in \autoref{img:midiffusion}.

\noindent
Moreover, Arena 5.0 introduces the capability to generate specific scenarios tailored to diverse environmental contexts such as emergency situations, normal operating hours, or unexpected scenarios like children in a warehouse or warehouse robots in a hospital setting. This feature also incorporates random behaviors—humans interacting unexpectedly with robots, such as running towards, playing with, or obstructing them—to ensure a wide variety of situations. This versatility is crucial for training and testing algorithms to enhance their robustness against unforeseen events.
\\\noindent
Exemplary scenarios across all three worlds are illustrated in \autoref{img:worlds}.
Within the graphical user interface (GUI), users have the option to select one of these predefined scenarios or customize their parameters to simulate different conditions. This flexibility allows researchers and developers to rigorously test and refine their navigation algorithms under controlled yet varied and realistic conditions.

\begin{figure*}[!htp]
    \centering
    \includegraphics[width=0.99\linewidth]{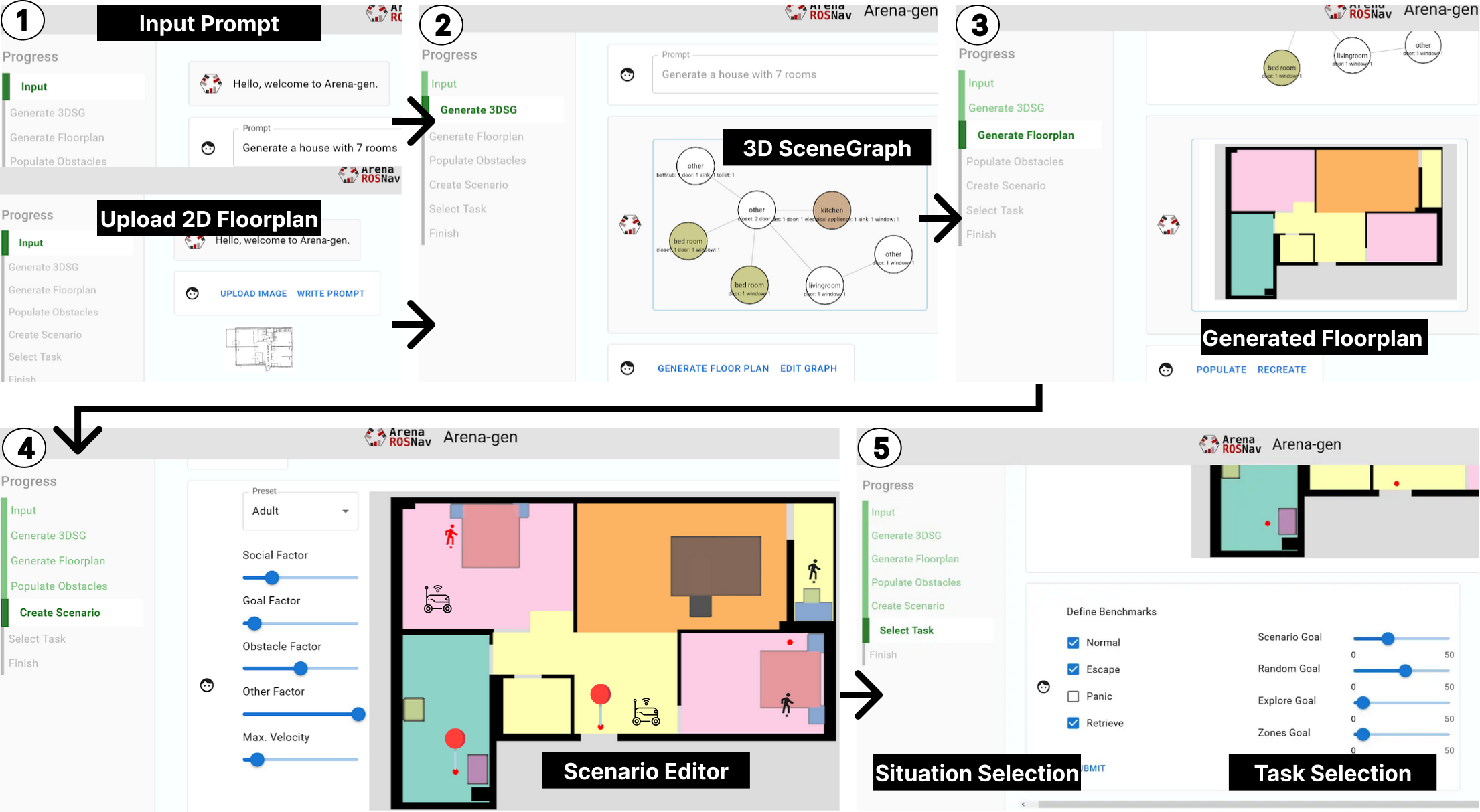}
    \caption{
    \emph{User Story of Arena-Gen Web App}:
    We provide a web frontend for our Arena-Gen module, which provides a convenient way for users to create synthetic worlds in an interactive chat. The user is (1) prompted to either write an input prompt or upload a floor plan which is converted into a (2) 3D Scene Graph that can be edited by the user (changing room layout, adjusting obstacles). When satisfied with the resulting 3DSG, the user can (3) generate floorplans based on the 3DSG until a suitable layout is achieved, then the population stage places the desired objects into sound positions within the room. Once the user is satisfied with the world generated in step (3), a default scenario is machine-created and then edited by the user (4). In the final step, (5) benchmarking tasks for this world can be enabled and weighted based on possible world situations. Finally, the world and configurations can be downloaded or exported to Arena directly. Overall, the user has a compact web interface that combines access to the various technologies used across all pipeline stages.  All functionalities are exposed through a stateless HTTP backend, with a central \texttt{/pipeline} backend that takes a text prompt and automatically passes through all following steps, for use in our simple CLI client.
    }
    \label{img:gen-userstory}
\end{figure*}

\subsection{Benchmarking and Dataset Acquisition Modules}
\noindent We have incorporated a diverse array of sensors into our platform, along with modules specifically designed to record, process, and plot data. Therefore, we implemented wrapper classes to integrate following sensors to work with all simulators including the newly integrated Isaac Gym simulator: 2D lidar, 3D lidar, RGB, RGBD camera, IMU, sonar sensor, foot contact sensor. These enhancements render the tool exceptionally well-suited for capturing large-scale, photorealistic datasets. moreover, by providing this set of diverse sensors, a wider variety of state of the art navigation approaches can be integrated and is not limited to 2D sensors like a majority of benchmarks. Furthermore, our automatic evaluation pipeline is designed to plot the results using a user friendly iPython notebook which is highly customizable. therefore, we also extended our metrics suite first presented in our previous work with more social metrics. A comprehensive overview of metric categories, metrics, and descriptions is listed in \autoref{tab:metrics}.

%% file: ieeeconf/content/4-evaluations.tex
\begin{figure*}[!h]
    \centering
    \includegraphics[width=0.99\linewidth]{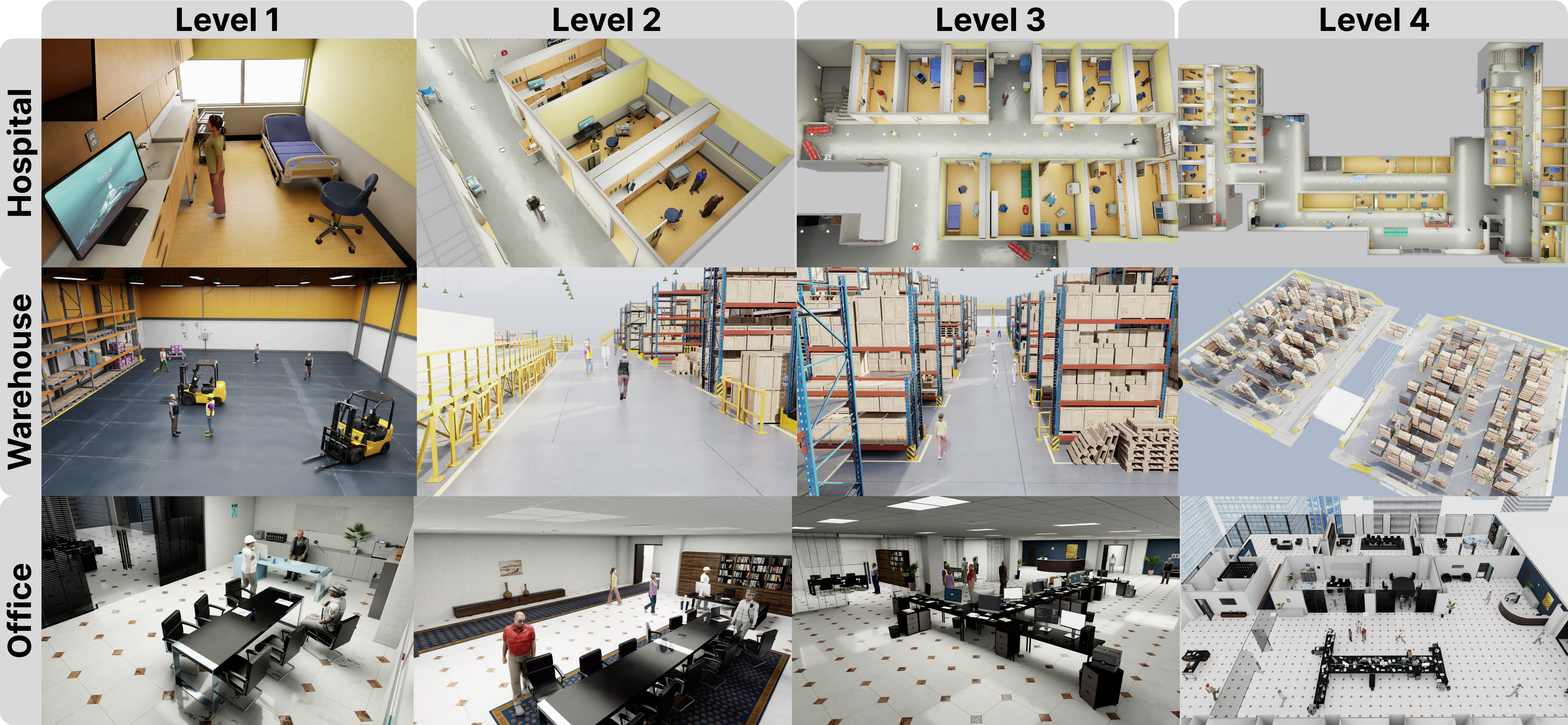}
    \caption{Example worlds generated using the Arena-Gen module for benchmarking and competition purposes. The worlds were generated using the text "generate 4 difficulty levels of a [hospital, residential, office] environment". The assets are automatically taken from the arena model database and pedestrians spawned with HuNavSim. Notably, a large variety of worlds for each environment type and level can be generated, e.g. 500 environments of hospital level 2. This feature aids quantitative benchmarking and in training new models. Users can also customize room layouts, pedestrian interactions with each other, asset placements, and specific situations using the Arena Architect GUI (shown in the supplementary video).}
    \label{img:worlds}
\end{figure*}
\section{Demonstrations and Exemplary Usage}
\noindent To demonstrate our platform and assess its functionalities, the following section will illustrate exemplary user stories using the provided web application for scenario generation. First, the web application is illustrated in \autoref{img:gen-userstory}, which showcases the user story to generate customized worlds and scenarios using our Arena-Gen integrated modules. Continuing the tradition established in our previous research, we conducted a study involving participants who were asked to install and test specific modules of the platform. In the following, we detail each of these steps.

\subsection{Creating Worlds and Scenarios Using the GUI}
\noindent Generating diverse environments is crucial for robustly benchmarking and testing approaches on robotic systems. We provide a web application in which the aforementioned Arena-Gen module is embedded. \autoref{img:gen-userstory} illustrates the step by step user story to generate a world and scenario. Note that the user can also decide to just create the world or scenario and doesn't necessarily need to provide pedestrians. First, the user can decide if upload a 2D floorplan, inputting a prompt or selecting a default prebuilt world (fig. step 1). Afterwards, the 3D scene graph is generated (2) and subsequently the floorplan (3), then the user can generate a scenario with pedestrian behavior or spawn more static obstacles using this generated floor plan. If the user decides to use a default world, the 2D floor plan is provided. In the next step, the user can add pedestrians and set a customized behavior, such as movement path, and other parameters for each pedestrian individually. To simplify this process, the user can also select default behavior situations such as routine activity in a hospital, or emergency situations where there is a default preset of parameters. Finally, the user sets a benchmark count for each task mode (from \autoref{tab:task_modes}) which sets the amount of task runs for the robot in each benchmark, ultimately represented in \autoref{lst:benchmark}. An exemplary benchmark configuration file is showcased in Listing 1.

\subsection{Platform for Competition and Benchmarking}
\begin{figure*}[!h]
    \centering
    \includegraphics[width=0.99\linewidth]{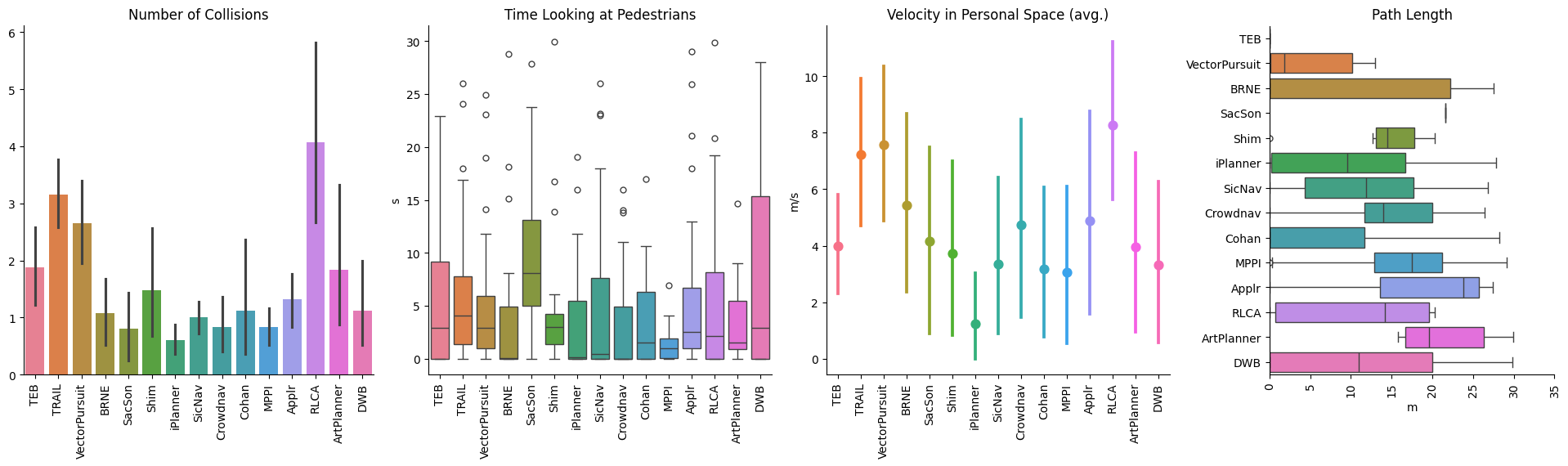}
    \caption{Example plots generated with Arena evaluation module for the conducted benchmark of all planners available in Arena 5.0, plotted on several social navigation metrics}
    \label{img:eval}
\end{figure*}

\noindent  Arena 5.0 significantly enhances evaluation capabilities by offering a suite of functions designed to facilitate the platform's use in unified benchmarking. The aforementioned Arena-Gen module has been extended to generate diverse worlds and scenarios, ensuring consistent and rigorous testing conditions, which is crucial for benchmarking (examples are illustrated in Figure \ref{img:worlds}).The APIs provided enable seamless integration of various methods and new planning algorithms. This functionality allows users to rigorously test their approaches in predefined or custom scenarios tailored to specific requirements and challenges. As a result of these advanced features, Arena 5.0 has been selected as the official hosting platform for competitions, including the forthcoming \href{https://socialnav2025.pages.dev/organization}{\color{blue}SocialNav2025 workshop}. 
\\\noindent
To demonstrate the platform's benchmarking abilities, we benchmarked state-of-the-art social navigation approaches on a variety of generated worlds. Therefore, we utilized our platform to generate worlds of different difficulty, more specifically hospital, office, and warehouse world each 4 levels are created. Subsequently, all planners~\cite{nav2-algos} listed in Table \autoref{tab:planners} are run for 15 runs on each world and the average value of each metric is calculated.
\\\noindent
The results of the benchmark are illustrated in \autoref{img:eval}, where each planners metric is displayed in various exemplary plotting styles. we also provide data preprocessing methods to calculate and plot the results effortlessly using an iPython notebook.

\subsubsection{User Study}
\noindent We evaluated our platform through a user study with 20 participants from universities in Germany, the US, Singapore, Japan, and Korea, representing diverse levels of robotics expertise. The group included individuals who had previously used Arena, social navigation researchers with no prior Arena experience, and students in computer science or engineering from multiple universities. Participants installed and interacted with the platform by completing tasks such as launching various task modes, initiating training runs, and testing different planners on multiple robots. They then filled out a questionnaire that assessed the platform’s strengths, weaknesses, overall impressions, suggestions for improvement, and ease of use, as well as their prior experience with Arena and background in robotics.
All responses and information about the participants occupation and level of expertise are publicly available on \href{https://drive.google.com/drive/folders/1lPJrvgxsfZyrIH3be6EFNX4R7y3PkL3J?usp=sharing}{\color{blue} Google Drive}. 
\\\noindent
The feedback primarily highlighted positive advancements in Arena 5.0 compared to earlier versions. Participants praised the new Isaac Gym integration and simplified wrappers, noting that they had previously avoided Isaac Sim due to limited documentation and complex setup. With Arena’s wrappers, many were able to quickly generate scenarios and benchmark approaches in Isaac Sim.
\\\noindent
Users also expressed favorable opinions about the web-based interface, stating that it is more intuitive and appealing than the previous JSON-based configuration workflow. The Arena-Gen modules were similarly well-received, with participants confirming that they effectively generate a variety of environments. However, some suggested expanding the current Generative AI pipeline beyond the three indoor environments (residential, office, and hospital) to include settings such as warehouses.
\\\noindent
Two participants reported occasional stability concerns, particularly longer wait times for world generation prompts, and only a few instances involved incorrect floorplans resulting in unrealistic 3D worlds. Additionally, several users cited an increased demand for computational power due to the integration of Isaac Sim and questioned whether earlier simulators would remain supported. In this regard, Arena’s modular and simulator-agnostic architecture ensures continued compatibility across multiple simulation tools.

%% file: ieeeconf/content/5-conclusion.tex
\section{Limitations}
Despite the potential and large variety of functionality this version offers, there are still limitations that are currently to be tackled.  The primary limitation of Arena~5.0 is its increased computational requirements: photorealistic simulation incurs higher overhead and latency than lighter‑weight simulators. However, we observe that performance scales more favorably with environment complexity and agent count, and that certain sensors are less computationally intensive to simulate than in Gazebo or Unity. Arena’s abstraction layer enables users to select the simulator best suited to their hardware and application needs. Human motion visualization remains dependent on pre‑existing animation assets, and creating new animations requires specialized 3D modeling expertise; we plan to address this by integrating motion‑capture transfer methods into the Arena ecosystem. Furthermore, world generation is currently limited to three environment types (office, hospital, and warehouse), and expanding this coverage via globally sourced datasets is a priority for future work. While no simulation can perfectly replicate reality, Arena~5.0 represents a substantive advance toward narrowing the sim‑to‑real gap.
\section{Conclusion}
\noindent 
In this paper, we introduce Arena 5.0, a substantial enhancement over our previous works Arena 1.0 \cite{kastner2021arena}, Arena 2.0 \cite{arena2}, Arena 3.0 \cite{arena3}, and Arena 4.0 \cite{arena4}, designed to support the dynamic generation of 3D environments for training, testing, and benchmarking social navigation approaches. A key upgrade in this version is the full integration of Isaac Gym, equipped with intuitive user interfaces and an improved world and scenario generation process. These enhancements simplify and enable the development and benchmarking of social navigation approaches within a unified platform.
\\\noindent
Additionally, Arena 5.0 is completely open-source, providing transparency and extensive customization options through accessible APIs that allow users to integrate their own planners and extend the platform's capabilities. The platform also includes comprehensive benchmarking tools and has incorporated the Navigation2~\cite{nav2} (Nav2) stack, along with several state-of-the-art planners and the most commonly used robotic platforms.
\\\noindent
With its user-friendly graphical interfaces and faster processing times, Arena 5.0 has demonstrated significant improvements in functionalities and user experience. Future developments aim to further expand the platform’s functionalities by integrating more advanced planning approaches, extend the world generation capabilities to include outdoor worlds and weather conditions, and multi-agent reinforcement learning approaches. The forthcoming release of Arena-Web~\cite{kastner2023arenaweb} will replicate most of the platform's features in a web-based application, enhancing its accessibility and ease of use for a global audience.

%% file: ieeeconf/content/6-appendix.tex
\section{Appendix}

\begin{table}[!h]
    \renewcommand{\arraystretch}{1.5}
    \scriptsize %
    \centering
    \caption{Overview of the robot and planner suites with descriptions of the planners and indications on which robot platform they can be deployed. Costmap planners are fully integrated with the nav2 framework and accept any number and types of sensors.}
    \begin{tabular}{m{1.0cm}m{0.8cm}m{1.0cm}m{4.2cm}}
    \hline 
    Planner & Type & Input & Description \\
    \hline\hline 
    Applr \cite{xiao2020appld} & Hybrid & 2D Lidar & A hybrid planner combining different approaches for adaptive planning. \\
    \hline 
    Cohan \cite{singamaneni2021human} & Classic & 2D Lidar & A traditional planner focusing on human-aware navigation strategies. \\
    \hline 
    Dragon \cite{xiao2022autonomous} & Hybrid & 2D Lidar & A hybrid navigation system designed for dynamic environments. \\
    \hline
    DWB \cite{khatib1986real} & Classic & Costmap & Modified version of the DWA Dynamic Window Approach, for ROS2 \\
    \hline 
    TEB \cite{rosmann2015timed} & Classic & Costmap & Timed Elastic Bands, optimizing a global path by considering kinematic and dynamic constraints. \\
    \hline
    Graceful \cite{park2011smooth} & Classic & Costmap & A Smooth Control Law for Graceful Motion of Differential Wheeled Mobile Robots in 2D Environment. \\
    \hline 
    MPPI \cite{williams2016aggressive} & Classic & Costmap &This is a predictive controller (local trajectory planner) that implements the Model Predictive Path Integral (MPPI) algorithm to track a path with adaptive collision avoidance. \\
    \hline
   SacSon \cite{xiao2022motion} & Learning & 2D Lidar & A hybrid planner leveraging both learned and heuristic components for effective navigation. \\
    \hline
    RLCA \cite{long2018towards} & Learning & 2D Lidar & Reinforcement Learning Collision Avoidance, using RL for dynamic obstacle avoidance. \\
    \hline
    ROSNavRL (OURS) & Learning & 2D Lidar & A learning-based approach trained on Arena 2.0 with Reinforcement Learning. \\
    \hline
    TRAIL \cite{xiao2022autonomous} & Learning & 2D Lidar & A learning-based planner focusing on trail navigation in unstructured environments. \\
    \hline
    Crowdnav \cite{chen2019crowd} & Learning & 2D Lidar & Focuses on navigating safely and efficiently in crowded environments using machine learning techniques. \\
    \hline
    SicNav \cite{sicnav} & Learning & Pedestrian & Socially Aware Reinforcement Learning, emphasizing social norms in navigation. \\
    \hline
    BRNE \cite{muchen23} & Game theory & Pedestrian & Utilizes game theory for decision-making in complex scenarios, suitable for quadruped robots. \\
    \hline
    iPlanner \cite{yangiplanner23} & Learning & RGBD & A learning-based planner for complex dynamic situations, suitable for quadruped robots. \\
    \hline
    ArtPlanner \cite{wellhausen2021rough} & RGB, RGBD & Quadruped & A learning-based planner optimized for articulated, quadruped robots. \\
    \hline
    RPPC~\cite{planner:rppc} & ROS2 & Costmap & implements a variant on the pure pursuit algorithm to track a path. This variant we call the Regulated Pure Pursuit Algorithm, due to its additional regulation terms on collision and linear speed.. \\
    \hline
    Shim~\cite{nav2-algos} & ROS2 & Costmap & The Rotation Shim Controller stands between the controller server and the main controller plugin to implement a specific behavior often troublesome for other algorithms. \\
    \hline
    Vector Pursuit~\cite{planner:vectorpursuit} & ROS2 & Costmap & Leverages Screw Theory to achieve accurate path tracking and comes with active collision detection. \\
    \hline
    \end{tabular}
    
    \label{tab:planners}
\end{table}

\begin{table}[!ht]
\centering
	\setlength{\tabcolsep}{0.2pt}
	\renewcommand{\arraystretch}{0.5}
		\caption{Overview of evaluation metrics}
	\begin{tabular}{lcp{3.5cm}}
		\hline \\
		Metric  &Unit & Explanation    \\
        \hline \\
        \rule{0pt}{2ex} \\
        \multicolumn{3}{c}{\footnotesize Performance} \\
        \hline \\
		Success Rate & \%        & Runs with < 2 collisions          \\ 
		Collision & -        & Total number of collisions\\
		Time to reach goal & [$s$] & Time required to reach the goal   \\ 
		Path Length  & [$m$]    & Path length in m           \\ 
		Velocity (avg.)     & [$\frac{m}{s}$]  & Velocity of the robot \\
		Acceleration (avg.) & [$\frac{m}{s^2}$] & Acceleration of the robot \\
		Movement Jerk & [$\frac{m}{s^3}$] & Derivation of Acceleration \\ 
		Curvature & [$m$] &  Degree of trajectory changes\\
		Angle over length & $[\frac{rad}{m}]$ & Curvature over the path-length \\
		Roughness & -                 & Quantifies trajectory smoothness \\ 
        Idling Time & $s$ & Time spent without moving \\ 
        Path Efficiency & $-$ & Ratio between the distance between two waypoints and the length of the agent’s actual path between those
        points \\ 
        \rule{0pt}{2ex} \\
        \multicolumn{3}{c}{\footnotesize Naturalness} \\
        \hline \\
        Average Displacement Error & $[m]$ & Average $L_2$ distance between the predicted trajectory and the human data \\
        Final Displacement Error & $[m]$ & Distance between the final destination in the prediction and the human data at the same
        time step \\
        Asymmetric Dynamic Time Warping & $[-]$ & A trajectory measure that doesn’t require both trajectories to have the same length \\
        \rule{0pt}{2ex} \\
        Path Irregularity  & $[-]$ & The amount of unnecessary turning over the whole path \\
        Topological Complexity  & $[-]$ & Measures path entanglement to quantify encounters \\
        \rule{0pt}{2ex} \\
        \multicolumn{3}{c}{\footnotesize Social Metrics} \\
        \hline \\
        Time seen by pedestrians & $[s]$ & Time robot was in pedestrians' field of view \\
        Time in personal space & $[s]$ & Time the robot spent in private zone \\
        \rule{0pt}{2ex} \\
        \multicolumn{3}{c}{\footnotesize Human Discomfort} \\
        \hline \\
        Minimum distance to human & $[m]$ & Minimum distance to any human during path \\
        Time in intimate space & \% & Percentage of time the robot spent in intimate space \\
        Max. speed in intimate space & $[\frac{m}{s}]$ & Maximum speed the robot had within intimate space \\
        Time facing pedestrians & $[s]$ & Time robot front faced pedestrians \\
        Movement towards pedestrians & \% & Percentage of path spent moving straight towards a pedestrian \\
		\hline
	\end{tabular}
	\label{tab:metrics}
\end{table}

\begin{table}[!ht]
    \centering
    \setlength{\tabcolsep}{0.2pt}
    \renewcommand{\arraystretch}{0.5}
    \caption{3dsg\_from\_text training parameters}
    \label{tab:llm_training}
    \begin{tabularx}{.8\linewidth}{XX}
            \hline \\
            \textbf{Parameter} & \textbf{Value} \\
            \hline \\
            Dataset Size & $\approx 60000$ \\
            Batch Size & $16$ \\
            Epochs & $10$ \\
            Optimizer & AdamW \\
            Learning Rate & $3\cdot10^{-5}$ \\
            Warmup Steps & $500$ \\
            Weight Decay & $0.01$ \\
            Dropout Rate & $0.1$ \\
            Gradient Accumulation & $4$ \\
            \hline \\
    \end{tabularx}
\end{table}

\begin{table}[!ht]
    \centering
    \setlength{\tabcolsep}{2pt}
    \renewcommand{\arraystretch}{0.5}
    \caption{\emph{3dsg\_from\_text dataset example}: }
    \label{tab:dataset}
    \begin{tabularx}{.99\linewidth}{lX}
            \hline \\
            \textbf{Prompt} & \makecell[l]{Generate a house with 3 rooms. \\ Add a toilet to the bathroom} \\
            \rule{0pt}{1ex} \\   
            \textbf{Room Output} & \makecell[l]{rooms=['Living Room', 'Bathroom', 'Bedroom'], \\connections=[(0,1),(0,2)]} \\
             \rule{0pt}{1ex} \\   
            \textbf{Object Output} & assets=[(1, 'Toilet')] \\
            \rule{0pt}{1ex} \\   
            \hline \\
    \end{tabularx}
\end{table}

\begin{table}
    \caption{Overview of all task modes.}
    \centering
    \scriptsize %
    \setlength{\tabcolsep}{2pt} %
    \renewcommand{\arraystretch}{1.1} %
    \begin{tabular}{m{1.5cm}m{1.5cm}m{5cm}}
        \hline
        \textbf{Parameter} & \textbf{Task modes} & \textbf{Explanation} \\ 
        \hline\hline
        Obstacle (tm\_obstacles) & Scenario & Loads obstacles described in scenario JSON file. \\ 
                                 & Random & Loads random obstacles defined in task\_manager.yaml file. \\
                                 & Parametrized & Loads random obstacles for each model specified in XML file. \\
                                 & Zones & Pedestrians are created with a role and waypoints are placed in semantically related zones. \\
                                 & Environment & Empty space in the world is filled with repeated obstacle clusters from a list of cluster definitions (classic world generation). \\
        \hline
        Robots (tm\_robots) & Scenario & Set start and goal for robot described in scenario JSON file. \\
                            & Random & Set random start and goal for robot. \\
                            & Guided & Changes RViz tool (2D Nav Goal) to set cyclic waypoints for robots and (2D pose estimate: reset robot position and sequence of waypoints). \\
                            & Explore & Obstacles are never reset and the robot receives random goals. \\
                            & Zones & Robots have an assigned role and receive tasks in world zones based on the semantics of their role. \\
        \hline
        Modules (tm\_modules) & Staged & Allows to switch between stages defined in training curriculum file. \\
                              & Dynamic & Creates a dynamic map. \\
                              & Benchmark & The user can provide a configuration to stage different task modes together. The benchmark configuration acts as a challenge or benchmark that other user can try and assess their planners. \\ 
        \hline
    \end{tabular}
    \label{tab:task_modes}
\end{table}

\begin{lstlisting}[
    float,
    caption={\emph{Example Benchmark Definition} for a care robot in a hospital environment. A scenario is evaluated 5 times, then 25 random tasks are completed.},
    label={lst:benchmark},
    basicstyle=\linespread{1}\tiny,
]
stages:
- config:
    SCENARIO:
      file: 1.json
  episodes: 5
  map: arena_hospital
  name: arena_hospital_1_cob4
  robot: cob4
  tm_obstacles: scenario
  tm_robots: scenario
- config:
    RANDOM:
      dynamic:
        min: 3
        max: 15
  episodes: 25
  map: arena_hospital
  name: arena_hospital_random_cob4
  robot: cob4
  tm_obstacles: random
  tm_robots: random
\end{lstlisting}